\documentclass[10pt,twocolumn,letterpaper]{article}

\usepackage{cvpr,times,epsfig,graphicx,amsmath,amssymb,microtype,subfig,multirow,nicefrac,amsfonts,mathtools,array,booktabs,bm,enumitem,algorithm,algorithmic,url}
\usepackage[pagebackref=true,breaklinks=true,letterpaper=true,colorlinks,bookmarks=false]{hyperref}
\usepackage[para,online,flushleft]{threeparttable}

\newcommand{\mc}{\mathcal}
\newcommand{\op}{\operatorname}
\newcommand{\lt}{\left}
\newcommand{\rt}{\right}
\newcommand{\ua}{$\uparrow$}
\newcommand{\da}{$\downarrow$}

\newcommand{\tb}{\textbf}

\newcommand{\grey}[1]{\textcolor[rgb]{0.618,0.618,0.618}{#1}}

\setlist[itemize]{itemsep=0pt, leftmargin=5mm}
\setlist[description]{itemsep=0pt, leftmargin=5mm}
\setlist[enumerate]{label=(\roman*), itemsep=0pt, leftmargin=7mm}

\cvprfinalcopy 

\def\cvprPaperID{648} 

\ifcvprfinal\fi

\begin{document}

\title{Tracking by Animation:\\Unsupervised Learning of Multi-Object Attentive Trackers}
\author{%
	Zhen He$^{1,2,3}$\thanks{Correspondence to Zhen He (email: {\tt hezhen.cs@gmail.com}).}\qquad Jian Li$^{2}$\qquad Daxue Liu$^{2}$\qquad Hangen He$^{2}$\qquad David Barber$^{3,4}$\\\\
	$^{1}$Academy of Military Medical Sciences\\
	$^{2}$National University of Defense Technology\\
	$^{3}$University College London\\
	$^{4}$The Alan Turing Institute%
}
\maketitle
\thispagestyle{empty}

\begin{abstract}
Online Multi-Object Tracking (MOT) from videos is a challenging computer vision task which has been extensively studied for decades.
Most of the existing MOT algorithms are based on the Tracking-by-Detection (TBD) paradigm combined with popular machine learning approaches which largely reduce the human effort to tune algorithm parameters.
However, the commonly used supervised learning approaches require the labeled data (e.g., bounding boxes), which is expensive for videos.
Also, the TBD framework is usually suboptimal since it is not end-to-end, i.e., it considers the task as detection and tracking, but not jointly.
To achieve both label-free and end-to-end learning of MOT, we propose a \emph{Tracking-by-Animation} framework, where a differentiable neural model first tracks objects from input frames and then animates these objects into reconstructed frames.
Learning is then driven by the reconstruction error through backpropagation.
We further propose a \emph{Reprioritized Attentive Tracking} to improve the robustness of data association.
Experiments conducted on both synthetic and real video datasets show the potential of the proposed model.
Our project page is publicly available
at: \url{https://github.com/zhen-he/tracking-by-animation}
\end{abstract}

\section{Introduction}

We consider the problem of online 2D multi-object tracking from videos.
Given the historical input frames, the goal is to extract a set of 2D object bounding boxes from the current input frame.
Each bounding box should have an one-to-one correspondence to an object and thus should not change its identity across different frames.

MOT is a challenging task since one must deal with:
(i) unknown number of objects, which requires the tracker to be correctly reinitialized/terminated when the object appears/disappears;
(ii) frequent object occlusions, which require the tracker to reason about the depth relationship among objects;
(iii) abrupt pose (e.g., rotation, scale, and position), shape, and appearance changes for the same object, or 
similar properties across different objects,
both of which make data association hard;
(iv) background noises (e.g., illumination changes and shadows), which can mislead tracking.

To overcome the above issues, one can seek to use expressive features, or improve the robustness of data association.
E.g., in the predominant Tracking-by-Detection (TBD) paradigm \cite{andriluka2008people,henriques2012exploiting,breitenstein2009robust,breitenstein2011online}, 
well-performed object detectors are first applied to extract object features (e.g., potential bounding boxes) from each input frame,
then appropriate matching algorithms are employed to associate these candidates of different frames, forming object trajectories.
To reduce the human effort to manually tune parameters for object detectors or matching algorithms, many machine learning approaches are integrated into the TBD framework and have largely improved the performance \cite{xiang2015learning,schulter2017deep,sadeghian2017tracking,milan2017online}.
However, most of these approaches are based on supervised learning, while manually labeling the video data is very time-consuming.
Also, the TBD framework does not consider the feature extraction and data association jointly, i.e., it is not end-to-end, thereby usually leading to suboptimal solutions.

In this paper, we propose a novel framework to achieve both label-free and end-to-end learning for MOT tasks.
In summary, we make the following contributions:
\begin{itemize}
	\item We propose a \emph{Tracking-by-Animation (TBA)} framework, where 
	a differentiable neural model first tracks objects from input frames and then animates these objects into reconstructed frames.
	Learning is then driven by the reconstruction error through backpropagation.
	\item We propose a \emph{Reprioritized Attentive Tracking (RAT)} to 
	mitigate overfitting and disrupted tracking, 
	improving the robustness of data association.
	\item We evaluate our model on two synthetic datasets (MNIST-MOT and Sprites-MOT) and one real dataset (DukeMTMC \cite{ristani2016performance}), showing its potential.
\end{itemize}

\section{Tracking by Animation}\label{sec_tba}

Our TBA framework consists of four components:
(i)   a \emph{feature extractor} that extracts input features from each input frame;
(ii)  a \emph{tracker array} where each tracker receives input features, updates its state, and emits outputs representing the tracked object;
(iii) a \emph{renderer} (parameterless) that renders tracker outputs into a reconstructed frame; 
(iv)  a \emph{loss} that uses the reconstruction error to drive the learning of Components~(i) and (ii), both label-free and end-to-end.

\subsection{Feature Extractor}

To reduce the computation complexity when associating trackers to the current observation, we first use a neural network $\op{NN}^{feat}$, parameterized by $\bm{\theta}^{feat}$, as a feature extractor to \emph{compress} the input frame at each timestep $t\!\in\!\{1, 2, \ldots, T\}$:
\begin{align}
\bm{C}_{t} = \op{NN}^{feat}\lt(\bm{X}_{t}; \bm{\theta}^{feat}\rt) \label{eq_feat}
\end{align}
where $\bm{X}_t \!\in \! [0,1]^{H \!\times \!W \!\times\! D}$ is the input frame of height $H$, width $W$, and channel size $D$, and $\bm{C}_{t} \!\in \! \mathbb{R}^{M \!\times \!N \!\times\! S}$ is the extracted input feature of height $M$, width $N$, and channel size $S$,
containing much fewer elements than $\bm{X}_t$.

\subsection{Tracker Array}\label{sec_tracker_array}

The tracker array comprises $I$ neural trackers indexed by $i\!\in\!\{1, 2, \ldots, I\}$ (thus $I$ is the maximum number of tracked objects).
Let $\bm{h}_{t,i}  \!\in \! \mathbb{R}^R$ be the state vector (vectors are assumed to be in row form throughout this paper) of Tracker $i$ at time $t$, and $\mc{H}_t\!=\!\{\bm{h}_{t,1}, \bm{h}_{t,2}, \ldots, \bm{h}_{t,I}\}$ be the set of all tracker states.
Tracking is performed by iterating over two stages:
\begin{enumerate}
	\item \tb{State Update}.
	The trackers first associate input features from $\bm{C}_{t}$ to update their states $\mc{H}_t$, through a neural network $\op{NN}^{upd}$ parameterized by $\bm{\theta}^{upd}$:
	\begin{align}
	\mc{H}_{t} = \op{NN}^{upd} \lt(\mc{H}_{t-1}, \bm{C}_{t}; \bm{\theta}^{upd}\rt) \label{eq_update}
	\end{align}
	Whilst it is straightforward to set $\op{NN}^{upd}$ as a Recurrent Neural Network (RNN) \cite{rumelhart1986learning,gers2000learning,cho2014learning} (with all variables vectorized), we introduce a novel RAT to model $\op{NN}^{upd}$ in order to increase the robustness of data association, which will be discussed in Sec.~\ref{sec_rat}.
	\item \tb{Output Generation}.
	Then, each tracker generates its output from $\bm{h}_{t,i}$ via a neural network $\op{NN}^{out}$ parameterized by $\bm{\theta}^{out}$:
	\begin{align}
	\mc{Y}_{t,i}  = \op{NN}^{out}\lt(\bm{h}_{t,i}; \bm{\theta}^{out}\rt) \label{eq_out}
	\end{align}
	where $\op{NN}^{out}$ is shared by all trackers, and the output $\mc{Y}_{t,i} = \lt\{y^c_{t,i}, \bm{y}^l_{t,i}, \bm{y}^p_{t,i}, \bm{Y}^s_{t,i}, \bm{Y}^a_{t,i}\rt\}$ is a \emph{mid-level} representation of objects on 2D image planes, including:
	\begin{description}
		\item[Confidence {$y^c_{t,i}\!\in\![0, 1]$}]
		~Probability of having captured an object, which can be thought
as a \emph{soft} sign of the trajectory validity (1/0 denotes valid/invalid). When time proceeds, an increase/decrease of $y^c_{t,i}$ can be thought as a
\emph{soft} initialization/termination of the trajectory.
		\item[Layer {$\bm{y}^l_{t,i}\!\in\!\{0,1\}^K$}]
		~One-hot encoding of the image layer possessed by the object.
		We consider each image to be composed of $K$ object layers and a background layer, where higher layer objects occlude lower layer objects and the background is the 0-th (lowest) layer.
		E.g., when $K\!=\!4$, $\bm{y}^l_{t,i}\!=\![0, 0, 1, 0]$ denotes the 3-rd layer.
		\item[Pose {$\bm{y}^p_{t,i}\!=\![\widehat{s}^x_{t,i}, \widehat{s}^y_{t,i}, \widehat{t}^x_{t,i}, \widehat{t}^y_{t,i}]\!\in\![-1,1]^4$}]
		~Normalized object pose for calculating the scale $[s^x_{t,i}, s^y_{t,i}] = [1 + \eta^{x} \widehat{s}^x_{t,i}, 1 + \eta^{y} \widehat{s}^y_{t,i}]$ and the translation $[t^x_{t,i}, t^y_{t,i}] = [\frac{W}{2}\widehat{t}^x_{t,i}, \frac{H}{2}\widehat{t}^y_{t,i}]$, where $\eta^{x},\eta^{y}>0$ are constants.
		\item[Shape {$\bm{Y}^s_{t,i}\!\in\!\{0,1\}^{U\!\times\!V\!\times1}$}]
		~Binary object shape mask with height $U$, width $V$, and channel size 1.
		\item[Appearance {$\bm{Y}^a_{t,i}\!\in\![0,1]^{U\!\times\!V\!\times\!D}$}]
		~Object appearance with height $U$, width $V$, and channel size $D$.
	\end{description}
	In the output layer of $\op{NN}^{out}$, $y^c_{t,i}$ and $\bm{Y}^a_{t,i}$ are generated by the sigmoid function, $\bm{y}^p_{t,i}$ is generated by the tanh function, and $\bm{y}^l_{t,i}$ and $\bm{Y}^s_{t,i}$ are sampled from the Categorical and Bernoulli distributions, respectively.
	As sampling is not differentiable, we use the Straight-Through Gumbel-Softmax estimator \cite{jang2017categorical} to reparameterize both distributions so that backpropagation can still be applied.
	
	The above-defined mid-level representation is not only flexible, but also can be directly used for input frame reconstruction, enforcing the output variables to be disentangled (as would be shown later).
	Note that through our experiments, we have found that the discreteness of $\bm{y}^l_{t,i}$ and $\bm{Y}^s_{t,i}$ is also very important for this disentanglement.
\end{enumerate}

\begin{figure*}[th]
	\centering
	\belowcaptionskip=-10pt
	\begin{minipage}{0.66\textwidth}
		\centering
		\includegraphics[height=63mm]{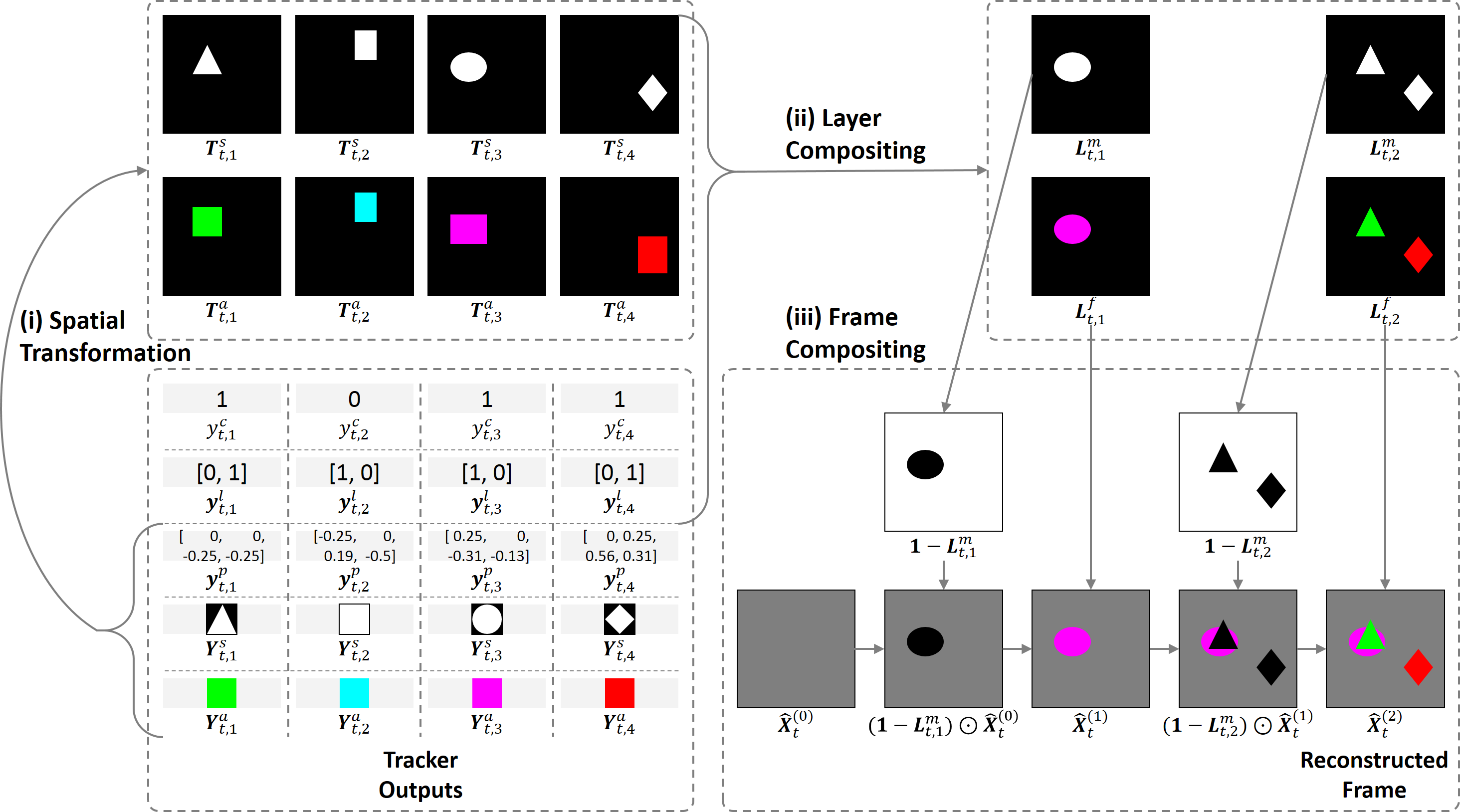}
		\caption{Illustration of the rendering process converting the tracker outputs into a reconstructed frame at time $t$, where the tracker number $I\!=\!4$ and the layer number $K\!=\!2$.}
		\label{fig-render}
	\end{minipage} \hfill
	\begin{minipage}{0.305\textwidth}
		\centering
		\includegraphics[height=63mm]{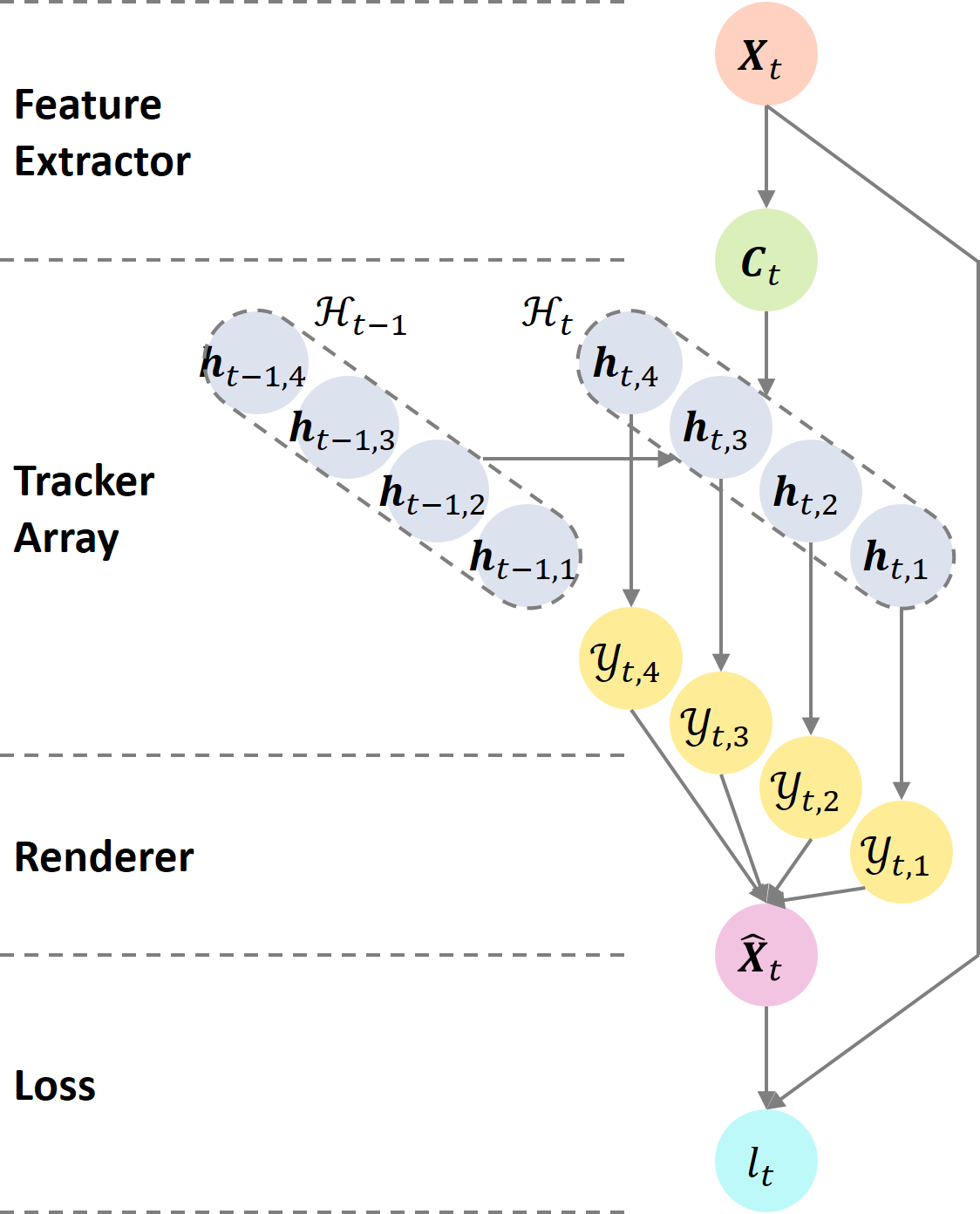}
		\caption{Overview of the TBA framework, where the tracker number $I\!=\!4$.\\}
		\label{fig-tba}
	\end{minipage}
\end{figure*}

\subsection{Renderer}

To define a training objective with only the tracker outputs but no training labels, 
we first use a differentiable renderer to convert all tracker outputs into reconstructed frames, and then minimize the reconstruction error through backpropagation. 
Note that we make the renderer both parameterless and deterministic so that correct tracker outputs can be encouraged in order to get correct reconstructions, enforcing the feature extractor and tracker array to learn to generate desired outputs.
The rendering process contains three stages:
\begin{enumerate}
	\item \tb{Spatial Transformation}.
	We first scale and shift $\bm{Y}_{t,i}^s$ and $\bm{Y}_{t,i}^a$ according to $\bm{y}_{t,i}^{p}$ via a Spatial Transformer Network (STN) \cite{jaderberg2015spatial}:
	\begin{align}
	\bm{T}_{t,i}^s & = \op{STN}\lt(\bm{Y}_{t,i}^{s}, \bm{y}_{t,i}^{p}\rt) \label{eq_render1} \\
	\bm{T}_{t,i}^a & = \op{STN}\lt(\bm{Y}_{t,i}^{a}, \bm{y}_{t,i}^{p}\rt)
	\end{align}
	where $ \bm{T}_{t,i}^s \! \in \! \{0, 1\}^{H \!\times \!W \!\times 1}$ and $\bm{T}_{t,i}^a \!\in \! [0, 1]^{H \!\times \!W \!\times\! D}$ are the spatially transformed shape and appearance, respectively.
	\item \tb{Layer Compositing}.
	Then, we synthesize $K$ image layers, where each layer can contain several objects. The $k$-th layer is composited by:
	\begin{align}
	\bm{L}_{t, k}^{m} &= \min \lt(1, \sum_{i} y^c_{t,i} y^l_{t,i,k} \bm{T}_{t,i}^s\rt) \label{eq_layer1} \\
	\bm{L}_{t, k}^{f} &= \sum_{i} y^c_{t,i} y^l_{t,i,k} \bm{T}_{t,i}^s \odot \bm{T}_{t,i}^a \label{eq_layer2}
	\end{align}
	where $\bm{L}_{t, k}^{m}\! \in \! [0, 1]^{H \!\times \!W \!\times 1}$ is the layer foreground mask, $\bm{L}_{t, k}^{f}\! \in \! [0, I]^{H \!\times \!W \!\times \! D}$ is the layer foreground, and $\odot$ is the element-wise multiplication which broadcasts its operands when they are in different sizes.
	\item \tb{Frame Compositing}.
	Finally, we iteratively reconstruct the input frame layer-by-layer, i.e., for $k\!=\! 1, 2, \ldots, K$:
	\begin{align}
	\widehat{\bm{X}}_{t}^{(k)}  = \lt(\bm{1} - \bm{L}_{t, k}^{m} \rt) \odot \widehat{\bm{X}}_{t}^{(k-1)} + \bm{L}_{t,k}^{f} \label{eq_render2}
	\end{align}
	where $\widehat{\bm{X}}_{t}^{(0)}$ is the extracted background, and  $\widehat{\bm{X}}_{t}^{(K)}$ is the final reconstruction. The whole rendering process is illustrated in Fig.~\ref{fig-render}, where $\eta^{x}\!=\!\eta^{y}\!=\!1$.
\end{enumerate}

Whilst the layer compositing can be parallelized by matrix operations, 
it cannot model occlusion since pixel values in overlapped object regions are simply added;
conversely, the frame compositing well-models occlusion, 
but the iteration process cannot be parallelized, consuming more time and memory.
Thus, we combine the two to both reduce the computation complexity and maintain the ability of occlusion modeling.
Our key insight is that though the number of occluded objects can be large, the occlusion \emph{depth} is usually small. Thus, occlusion can be modeled efficiently by using a small layer number $K$ (e.g., $K\!=\!3$), in which case each layer will be shared by several non-occluded objects.

\subsection{Loss}
To drive the learning of the feature extractor as well as the tracker array, we define a loss $l_t$ for each timestep:
\begin{align}
l_t = \op{MSE} \lt(\widehat{\bm{X}}_{t}, \bm{X}_{t} \rt) + \lambda \cdot \frac{1}{I}\sum_{i} s^x_{t,i}~s^y_{t,i} \label{eq_loss}
\end{align}
where, on the RHS, the first term is the reconstruction Mean Squared Error, and the second term, weighted by a 
constant $\lambda\!>\!0$, is the \emph{tightness} constraint penalizing large scales $[s^x_{t,i}, s^y_{t,i}]$ in order to make object bounding boxes more compact.
An overview of our TBA framework is shown in Fig.~\ref{fig-tba}.

\section{Reprioritized Attentive Tracking}\label{sec_rat}

In this section, we focus on designing the tracker state update network $\op{NN}^{upd}$ defined in (\ref{eq_update}).
Although $\op{NN}^{upd}$ can be naturally set as a single RNN as mentioned in Sec.~\ref{sec_tracker_array}, there can be two issues:
(i)  \emph{overfitting}, since there is no mechanism to capture the data regularity that similar patterns are usually shared by different objects;
(ii) \emph{disrupted tracking}, since there is no incentive to drive each tracker to associate its relevant input features.
Therefore, we propose the RAT, which
tackles Issue~(i) by modeling each tracker independently and sharing parameters for different trackers (this also reduces the parameter number and makes learning more scalable with the tracker number), and
tackles Issue~(ii) by utilizing attention to achieve explicit data association (Sec.~\ref{subsec_att}).
RAT also avoids \emph{conflicted tracking} by employing memories to allow tracker interaction (Sec.~\ref{subsec_mem}) and 
reprioritizing trackers to make data association more robust (Sec.~\ref{subsec_rep}), and
improves efficiency by adapting the computation time according to the number of objects presented in the scene~(Sec.~\ref{subsec_act}).

\subsection{Using Attention}\label{subsec_att}

To make Tracker $i$ explicitly associate its relevant input features from $\bm{C}_{t}$ to avoid disrupted tracking, we adopt a content-based addressing.
Firstly, the previous tracker state $\bm{h}_{t-1,i}$ is used to generate key variables $\bm{k}_{t,i}$ and  $\beta_{t,i}$:
\begin{align}
\lt\{\bm{k}_{t,i}, \widehat{\beta}_{t,i}\rt\} & = \op{Linear}\lt(\bm{h}_{t-1, i}; \bm{\theta}^{key}\rt) \label{eq_read1} \\
\beta_{t,i} & = 1 + \ln \lt(1 +\exp\lt({\widehat{\beta}_{t,i}}\rt)\rt)
\end{align}
where $\op{Linear}$ is the linear transformation parameterized by $\bm{\theta}^{key}$, $\bm{k}_{t,i}\!\in\!\mathbb{R}^S$ is the addressing key, and $\widehat{\beta}_{t,i}\!\in\!\mathbb{R}$ is the activation for the key strength $\beta_{t,i}\!\in\!(1, +\infty)$.
Then, $\bm{k}_{t,i}$ is used to match each feature vector in $\bm{C}_{t}$, denoted by $\bm{c}_{t,m,n}\!\in\!\mathbb{R}^S$ where $m\!\in\!\{1, 2, \ldots, M\}$ and $n\!\in\!\{1, 2, \ldots, N\}$, to get attention weights:
\begin{align}
W_{t,i,m,n} = \frac{\exp \Big(\beta_{t,i} \op{K}\lt( \bm{k}_{t,i}, \bm{c}_{t,m,n} \rt) \Big)}{\sum_{m',n'} \exp \Big(\beta_{t,i} \op{K}\lt( \bm{k}_{t,i}, \bm{c}_{t,m',n'} \rt) \Big)} \label{eq_attention}
\end{align}
where $\op{K}$ is the cosine similarity defined as $\op{K}\lt(\bm{p}, \bm{q}\rt)\!=\! \bm{p} \bm{q}^{\mathsf{T}} / \lt(\|\bm{p}\| \|\bm{q}\|\rt)$, and $W_{t,i,m,n}$ is an element of the attention weight $\bm{W}_{t,i}\!\in\![0,1]^{M\!\times\!N}$, satisfying $\sum_{m,n}\!W_{t,i,m,n}\!=\!1$.
Next, a read operation is defined as a weighted combination of all feature vectors of $\bm{C}_{t}$:
\begin{align}
&\bm{r}_{t,i} = \sum_{m,n} W_{t,i,m,n}~\bm{c}_{t,m,n} \label{eq_read2}
\end{align}
where $\bm{r}_{t,i}\!\in\!\mathbb{R}^S$ is the read vector, representing the associated input feature for Tracker $i$.
Finally, the tracker state is updated with an RNN parameterized by $\bm{\theta}^{rnn}$, taking $\bm{r}_{t,i}$ instead of $\bm{C}_{t}$ as its input feature:
\begin{align}
\bm{h}_{t, i} = \op{RNN} \lt(\bm{h}_{t-1, i}, \bm{r}_{t,i}; \bm{\theta}^{rnn}\rt) \label{eq_read3}
\end{align}

Whilst each tracker can now attentively access $\bm{C}_{t}$, it still cannot attentively access $\bm{X}_{t}$ if the receptive field of each feature vector $\bm{c}_{t,m,n}$ is too large.
In this case, it remains hard for the tracker to correctly associate an object from $\bm{X}_{t}$.
Therefore, we set the feature extractor $\op{NN}^{feat}$ as a Fully Convolutional Network (FCN) \cite{long2015fully,xu2015show,wang2015visual} purely consisting of convolution layers.
By designing the kernel size of each convolution/pooling layer, we can control the receptive field of $\bm{c}_{t,m,n}$ to be a local region on the image so that the tracker can also attentively access $\bm{X}_{t}$.
Moreover, parameter sharing in FCN captures the spatial regularity that similar patterns are shared by objects on different image locations.
As a local image region contains little information about the object translation $[t^x_{t,i}, t^y_{t,i}]$, we add this information by appending the 2D image coordinates as two additional channels to $\bm{X}_{t}$.

\subsection{Input as Memory}\label{subsec_mem}

To allow trackers to interact with each other to avoid conflicted tracking, at each timestep, we take the input feature $\bm{C}_t$ as an external memory through which trackers can pass messages.
Concretely, Let $\bm{C}_{t}^{(0)}\!=\!\bm{C}_{t}$ be the initial memory, we arrange trackers to sequentially read from and write to it, so that $\bm{C}_{t}^{(i)}$ records all messages written by the past $i$ trackers.
In the $i$-th iteration ($i\!=\! 1, 2, \ldots, I$), Tracker $i$ first reads from $\bm{C}_{t}^{(i-1)}$ to update its state $\bm{h}_{t, i}$ by using (\ref{eq_read1})--(\ref{eq_read3}) (where $\bm{C}_{t}$ is replaced by $\bm{C}_{t}^{(i-1)}$).
Then, an erase vector $\bm{e}_{t,i}\!\in\![0, 1]^S$ and a write vector $\bm{v}_{t,i}\!\in\!\mathbb{R}^S$ are emitted by:
\begin{align}
\lt\{\widehat{\bm{e}}_{t,i}, \bm{v}_{t,i} \rt\} &= \op{Linear}\lt(\bm{h}_{t, i}; \bm{\theta}^{wrt}\rt) \label{eq_write1} \\
\bm{e}_{t,i} &=  \op{sigmoid}\lt(\widehat{\bm{e}}_{t,i}\rt) \label{eq_write2} 
\end{align}
With the attention weight $\bm{W}_{t,i}$ produced by (\ref{eq_attention}), we then define a write operation, where each feature vector in the memory is modified as:
\begin{align}
\!\!\bm{c}_{t,m,n}^{(i)} = \lt(\bm{1}-W_{t,i,m,n} \bm{e}_{t,i}\rt) \odot \bm{c}_{t,m,n}^{(i-1)} + W_{t,i,m,n} \bm{v}_{t,i} \label{eq_write3}
\end{align}

Our tracker state update network defined in (\ref{eq_read1})--(\ref{eq_write3}) is inspired by the Neural Turing Machine \cite{graves2014neural,graves2016hybrid}.
Since trackers (controllers) interact through the external memory by using interface variables, they do not need to encode messages of other trackers into their own working memories~(i.e., states), making tracking more efficient.

\subsection{Reprioritizing Trackers}\label{subsec_rep}

Whilst memories are used for tracker interaction, it is hard for high-priority (small $i$) but low-confidence trackers to associate data correctly.
E.g., when the first tracker ($i\!=\!1$) is free ($y^c_{t-1,1}\!=\!0$), it is very likely for it to associate or, say, `steal' a tracked object from a succeeding tracker, since from the unmodified initial memory $\bm{C}_{t}^{(0)}$, all objects are equally chanced to be associated by a free tracker.

To avoid this situation, we first update high-confidence trackers so that features corresponding to the tracked objects can be firstly associated and modified.
Therefore, we define the priority $p_{t, i}\!\in\!\{1,2,\ldots,I\}$ of Tracker $i$ as its previous (at time $t\!-\!1$) confidence ranking (in descending order) instead of its index $i$, and then we can update Tracker $i$ in the $p_{t, i}$-th iteration to make data association more robust.

\subsection{Using Adaptive Computation Time}\label{subsec_act}

\begin{figure}[t]
	\centering
	\belowcaptionskip=-5pt
	\abovecaptionskip=5pt
	\includegraphics[height=38mm]{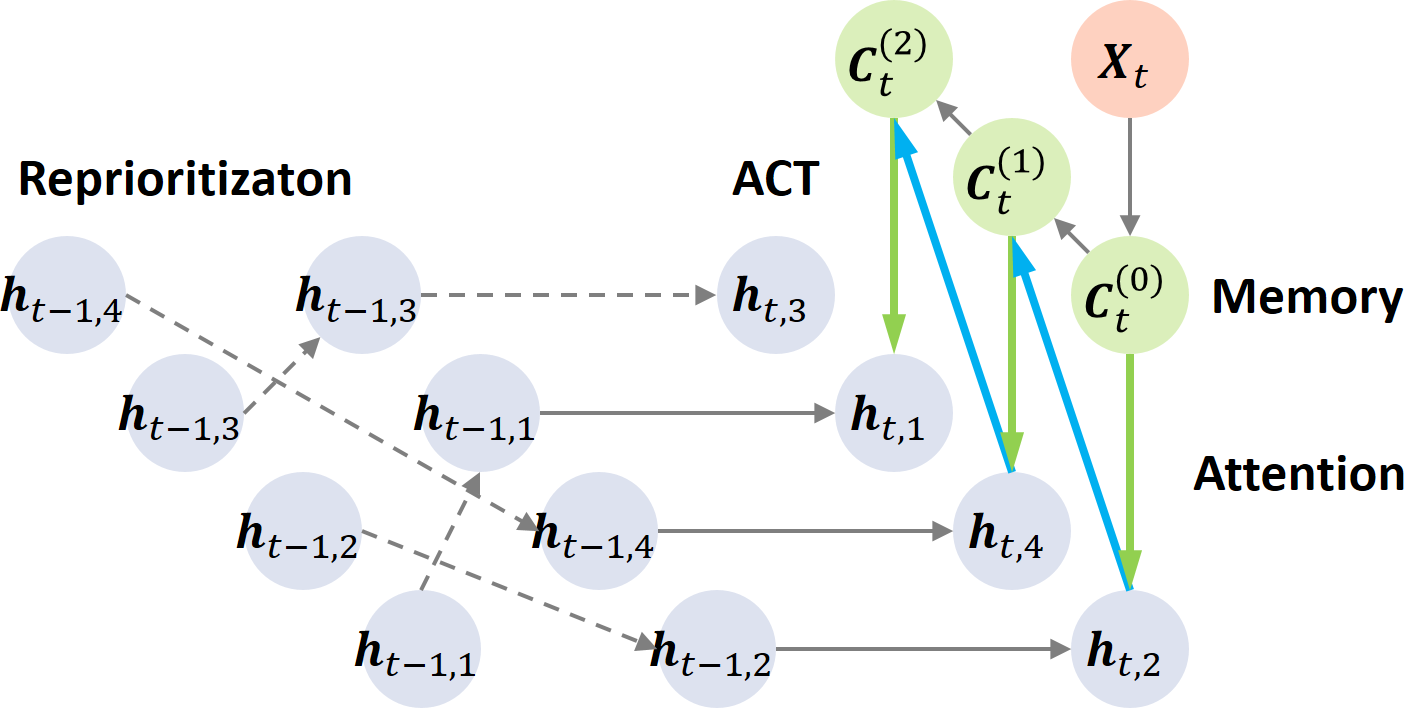}
	\caption{Illustration of the RAT with the tracker number $I\!=\!4$.
		Green/Blue bold lines denote attentive read/write operations on memory.
		Dashed arrows denote copy operations.
		At time~$t$, the iteration is performed by 3 times and terminated at Tracker~1.}
	\label{fig-rat}
\end{figure}

Since the object number varies with time and is usually less than the tracker number $I$ (assuming $I$ is set large enough), iterating over all trackers at every timestep is inefficient. 
To overcome this, we adapt the idea of Adaptive Computation Time (ACT) \cite{graves2016adaptive} to RAT.
At each timestep~$t$, we terminate the iteration at Tracker $i$ (also disable the write operation) once $y^c_{t-1,i}\!<\!0.5$ and $y^c_{t,i}\!<\!0.5$, in which case there are unlikely to be more tracked/new objects.
While for the remaining trackers, we do no use them to generate outputs.
An illustration of the RAT is shown in Fig.~\ref{fig-rat}.
The algorithm of the full TBA framework is presented in Fig.~\ref{fig-algo}.\vspace{-3pt}

\begin{figure}[ht]
	\small
	\centering
	\abovecaptionskip=2pt
	\belowcaptionskip=-8pt
	\arrayrulewidth = 1pt
	\begin{tabular}{m{80mm}}
		\hline\\[-8pt]
		\begin{algorithmic}[1]
			\STATE \grey{\# Initialization}
			\FOR{$i \gets 1$ \TO $I$}
			\STATE $\bm{h}_{0,i} \gets \bm{0}$
			\STATE $y_{0,i}^c \gets 0$
			\ENDFOR
			\STATE \grey{\# Forward pass}
			\FOR{$t \gets 1$ \TO $T$}
			\STATE \grey{\# (i) Feature extractor}
			\STATE extract $\bm{C}_t$ from $\bm{X}_t$, see (\ref{eq_feat})
			\STATE \grey{\# (ii) Tracker array}
			\STATE $\bm{C}_t^{(0)} \gets \bm{C}_t$
			\STATE use $y_{t-1,1}^c, y_{t-1,2}^c, \ldots, y_{t-1,I}^c$ to calculate $p_{t,1}, p_{t,2}, \ldots, p_{t,I}$ 
			\FOR{$j \gets 1$ \TO $I$}
			\STATE select the $i$-th tracker whose priority $p_{t,i} = j$
			\STATE use $\bm{h}_{t-1,i}$ and $\bm{C}_t^{(j-1)}$ to generate $\bm{W}_{t,i}$, see (\ref{eq_read1})--(\ref{eq_attention})\!\!\!
			\STATE read from $\bm{C}_t^{(j-1)}$ according to $\bm{W}_{t,i}$, and update $\bm{h}_{t-1,i}$ to $\bm{h}_{t,i}$, see (\ref{eq_read2}) and (\ref{eq_read3})
			\STATE use $\bm{h}_{t,i}$ to generate $\mc{Y}_{t,i}$, see (\ref{eq_out})
			\IF{$y_{t-1,i}<0.5$ \AND $y_{t,i}<0.5$}
			\STATE \tb{break} 
			\ENDIF
			\STATE write to $\bm{C}_t^{(j-1)}$ using $\bm{h}_{t,i}$ and $\bm{W}_{t,i}$, obtaining $\bm{C}_t^{(j)}$, see (\ref{eq_write1})--(\ref{eq_write3})
			\ENDFOR
			\STATE \grey{\# (iii) Renderer}
			\STATE use $\mc{Y}_{t,1}, \mc{Y}_{t,2}, \ldots, \mc{Y}_{t,I}$ to 
			render $\widehat{\bm{X}}_t$, see (\ref{eq_render1})--(\ref{eq_render2})
			\STATE \grey{\# (iv) Loss}
			\STATE calculate $l_t$, see (\ref{eq_loss})
			\ENDFOR
		\end{algorithmic}\\[-9pt]\hline\\[-7pt]
	\end{tabular}
	\caption{Algorithm of the TBA framework.}\label{fig-algo}
\end{figure}

\pagebreak

\section{Experiments}

The main purposes of our experiments are:
(i) investigating the importance of each component in our model, and
(ii) testing whether our model is applicable to real videos.
For Purpose (i), we create two synthetic datasets (MNIST-MOT and Sprites-MOT), and consider the following configurations:
\begin{description}
	\item[TBA] The full TBA model as described in Sec.~\ref{sec_tba} and Sec.~\ref{sec_rat}.
	\item[TBAc] TBA with constant computation time, by not using the ACT described in Sec.~\ref{subsec_act}.
	\item[TBAc-noOcc] TBAc without occlusion modeling, by setting the layer number $K\!=\!1$.
	\item[TBAc-noAtt] TBAc without attention, by reshaping the memory $\bm{C}_{t}$ into size $[1,1,MNS]$, in which case the attention weight degrades to a scalar ($\bm{W}_{t,i}\!=\!W_{t,i,1,1}\!=\!1$).
	\item[TBAc-noMem] TBAc without memories, by disabling the write operation defined in (\ref{eq_write1})--(\ref{eq_write3}).
	\item[TBAc-noRep] TBAc without the tracker reprioritization described in Sec.~\ref{subsec_rep}.
	\item[AIR] Our implementation of the `Attend, Infer, Repeat'~(AIR)~\cite{eslami2016attend} for qualitative evaluation, which is a probabilistic generative model that can be used to detect objects from individual images through inference.
\end{description}
Note that it is hard to set a supervised counterpart of our model for online MOT, since calculating the supervised loss with ground truth data is \textit{per se} an optimization problem which requires to access complete trajectories and thus is usually done offline \cite{schulter2017deep}.
For Purpose (ii), we evaluate TBA on the challenging DukeMTMC dataset \cite{ristani2016performance}, and compare it to the state-of-the-art methods.
In this paper, we only consider videos with static backgrounds $\widehat{\bm{X}}_t^{(0)}$, and use the IMBS algorithm \cite{bloisi2012independent} to extract them for input reconstruction.

Implementation details of our experiments are given in Appendix~\ref{app_implement}.
The MNIST-MOT experiment is reported in Appendix~\ref{app_mnist}.
The appendix can be downloaded from our project page.

\subsection{Sprites-MOT}

\begin{figure}[b]
	\centering
	\abovecaptionskip=5pt
	\vspace{-3pt}
	\includegraphics[width=0.4\textwidth]{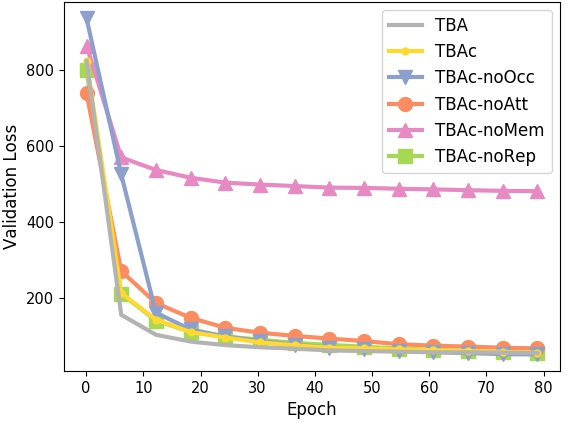}
	\caption{Training curves of different configurations on Sprites-MOT.}
	\label{fig-curve-sprite}
\end{figure}

\begin{figure*}[th]
	\centering
	\abovecaptionskip=5pt
	\belowcaptionskip=-3pt
	\includegraphics[width=0.994\textwidth]{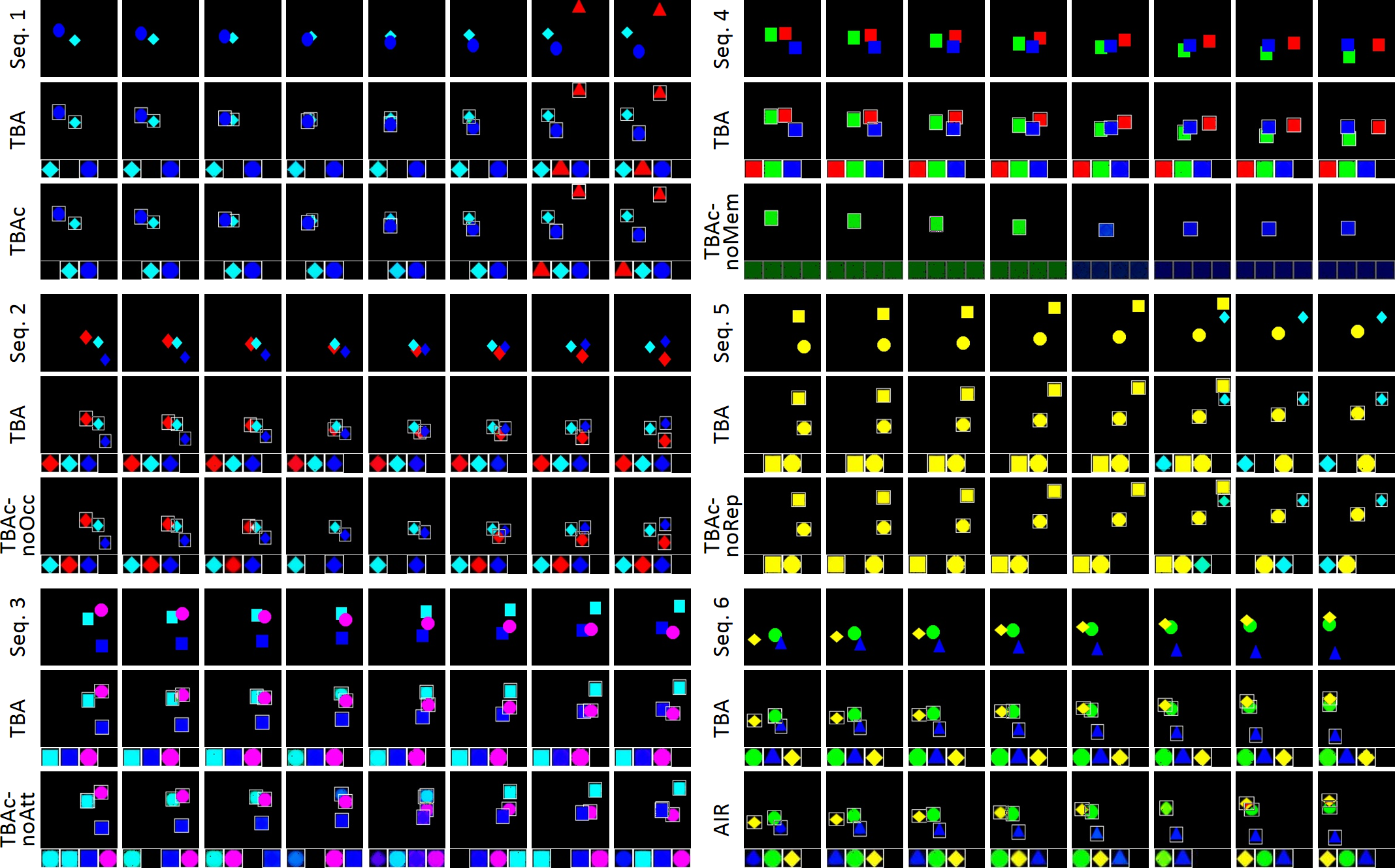}
	\caption{Qualitative results of different configurations on Sprites-MOT.
		For each configuration, we show the reconstructed frames (top) and the tracker outputs (bottom). 
		For each frame, tracker outputs from left to right correspond to tracker 1 to $I$ (here $I\!=\!4$), respectively.
		Each tracker output $\mc{Y}_{t,i}$ is visualized as $\lt(y^c_{t,i}~\bm{Y}_{t,i}^s \odot \bm{Y}_{t,i}^a\rt) \in [0,1]^{U\!\times\!V\!\times\!D}$.}
	\label{fig-quali-sprite}
\end{figure*}

\begin{table*}[th]
	\small
	\centering
	\tabcolsep=5pt
	\abovecaptionskip=0pt
	\caption{Tracking performances of different configurations on Sprites-MOT.}
	\begin{tabular}{ccccccccccccc}
		\toprule
		\tb{Configuration} &\tb{IDF1\ua}&\tb{IDP\ua}&\tb{IDR\ua}&\tb{MOTA\ua}&\tb{MOTP\ua}&\tb{FAF\da}&\tb{MT\ua}&\tb{ML\da}&\tb{FP\da}&\tb{FN\da}&\tb{IDS\da}&\tb{Frag\da}\\
		\midrule
		TBA         &  99.2  &  99.3  &  99.2  &  99.2  &  79.1  &  0.01  &  985  &    1  &    60  &     80  &    30  &    22   \\
		TBAc        &  99.0  &  99.2  &  98.9  &  99.1  &  78.8  &  0.01  &  981  &    0  &    72  &     83  &    36  &    29   \\
		TBAc-noOcc  &  93.3  &  93.9  &  92.7  &  98.5  &  77.9  &     0  &  969  &    0  &    48  &    227  &    64  &   105   \\
		TBAc-noAtt  &  43.2  &  41.4  &  45.1  &  52.6  &  78.6  &  0.19  &  982  &    0  & 1,862  &    198  & 8,425  &    89   \\
		TBAc-noMem  &     0  &    --  &     0  &     0  &    --  &     0  &    0  &  987  &     0  & 22,096  &     0  &     0   \\
		TBAc-noRep  &  93.0  &  92.5  &  93.6  &  96.9  &  78.8  &  0.02  &  978  &    0  &   232  &    185  &   267  &    94   \\
		\bottomrule
		\vspace{-22pt}
	\end{tabular}\label{tab-quanti-sprite}
\end{table*}

In this toy task, we aim to test whether our model can robustly handle occlusion and track the pose, shape, and appearance of the object that can appear/disappear from the scene, providing accurate and consistent bounding boxes.
\noindent Thus, we create a new Sprites-MOT dataset containing 2M frames,
where each frame is of size 128$\times$128$\times$3,
consisting of a black background and at most three moving sprites that can occlude each other.
Each sprite is randomly scaled from a 21$\times$21$\times$3 image patch 
with a random shape (circle/triangle/rectangle/diamond) and a random color (red/green/blue/yellow/magenta/cyan),
moves towards a random direction, and appears/disappears only once.
To solve this task, for TBA configurations 
we set the tracker number $I\!=\!4$ and layer number $K\!=\!3$.

Training curves are shown in Fig.~\ref{fig-curve-sprite}.
TBAc-noMem has the highest validation loss, indicating that it cannot well reconstruct the input frames,
while other configurations perform similarly and have significantly lower validation losses.
However, TBA converges the fastest, which we conjecture benefits from the regularization effect introduced by ACT.

To check the tracking performance, we compare TBA against other configurations on several sampled sequences, as shown in Fig.~\ref{fig-quali-sprite}.
We can see that TBA consistently performs well on all situations, where in Seq.~1 TBAc perform as well as TBA.
However, TBAc-noOcc fails to track objects from occluded patterns (in Frames 4 and 5 of Seq.~2, the red diamond is lost by Tracker 2).
We conjecture the reason is that adding values of occluded pixels into a single layer can result in high reconstruction errors, and thereby the model just learns to suppress tracker outputs when occlusion occurs.
Disrupted tracking frequently occurs on TBAc-noAtt which does not use attention explicitly (in Seq.~3, trackers frequently change their targets).
For TBAc-noMem, all trackers know nothing about each other and compete for a same object, resulting in identical tracking with low confidences.
For TBAc-noRep, free trackers incorrectly associate the objects tracked by the follow-up trackers.
Since AIR does not consider the temporal dependency of sequence data, it fails to track objects across different timesteps.

We further quantitatively evaluate different configurations using the standard CLEAR MOT metrics (Multi-Object Tracking Accuracy (MOTA), Multi-Object Tracking Precision (MOTP), etc.) \cite{bernardin2008evaluating} that count how often the tracker makes incorrect decisions, and 
the recently proposed ID metrics (Identification F-measure (IDF1), Identification Precision (IDP), and  Identification Recall (IDR)) \cite{ristani2016performance} that measure how long the tracker correctly tracks targets.
Note that we only consider tracker outputs $\mc{Y}_{t,i}$ with confidences $y^c_{t,i}\!>\!0.5$ and convert the corresponding poses $\bm{y}^p_{t,i}$ into object bounding boxes for evaluation.
Table~\ref{tab-quanti-sprite} reports the tracking performance.
Both TBA and TBAc gain good performances and TBA performs slightly better than TBAc.
For TBAc-noOcc, it has a significantly higher False Negative (FN) (227), ID Switch (IDS) (64), and Fragmentation (Frag) (105),
which is consistent with our conjecture from the qualitative results that using a single layer can sometimes suppress tracker outputs.
TBAc-noAtt performs poorly on most of the metrics, especially with a very high IDS of 8425 potentially caused by disrupted tracking.
Note that TBAc-noMem has no valid outputs as all tracker confidences are below 0.5.
Without tracker reprioritization, TBAc-noRep is less robust than TBA and TBAc, with a higher False Positive (FP) (232), FN (185), and IDS (267) that we conjecture are mainly caused by conflicted tracking.

\subsection{DukeMTMC}

To test whether our model can be applied to the real applications involving highly complex and time-varying data patterns, we evaluate the full TBA on the challenging DukeMTMC dataset \cite{ristani2016performance}.
It consists of 8 videos of resolution 1080$\times$1920,
with each split into 50/10/25 minutes long for training/test(hard)/test(easy).
The videos are taken from 8 fixed cameras recording movements of people on various places of Duke university campus at 60fps.
For TBA configurations, we set the tracker number $I\!=\!10$ and layer number $K\!=\!3$.
Input frames are down-sampled to 10fps and resized to 108$\times$192 to ease processing.
Since the hard test set contains very different people statistics from the training set, we only evaluate our model on the easy test set.

\begin{figure*}[th]
	\centering
	\abovecaptionskip=5pt
	\belowcaptionskip=-13pt
	\includegraphics[width=0.994\textwidth]{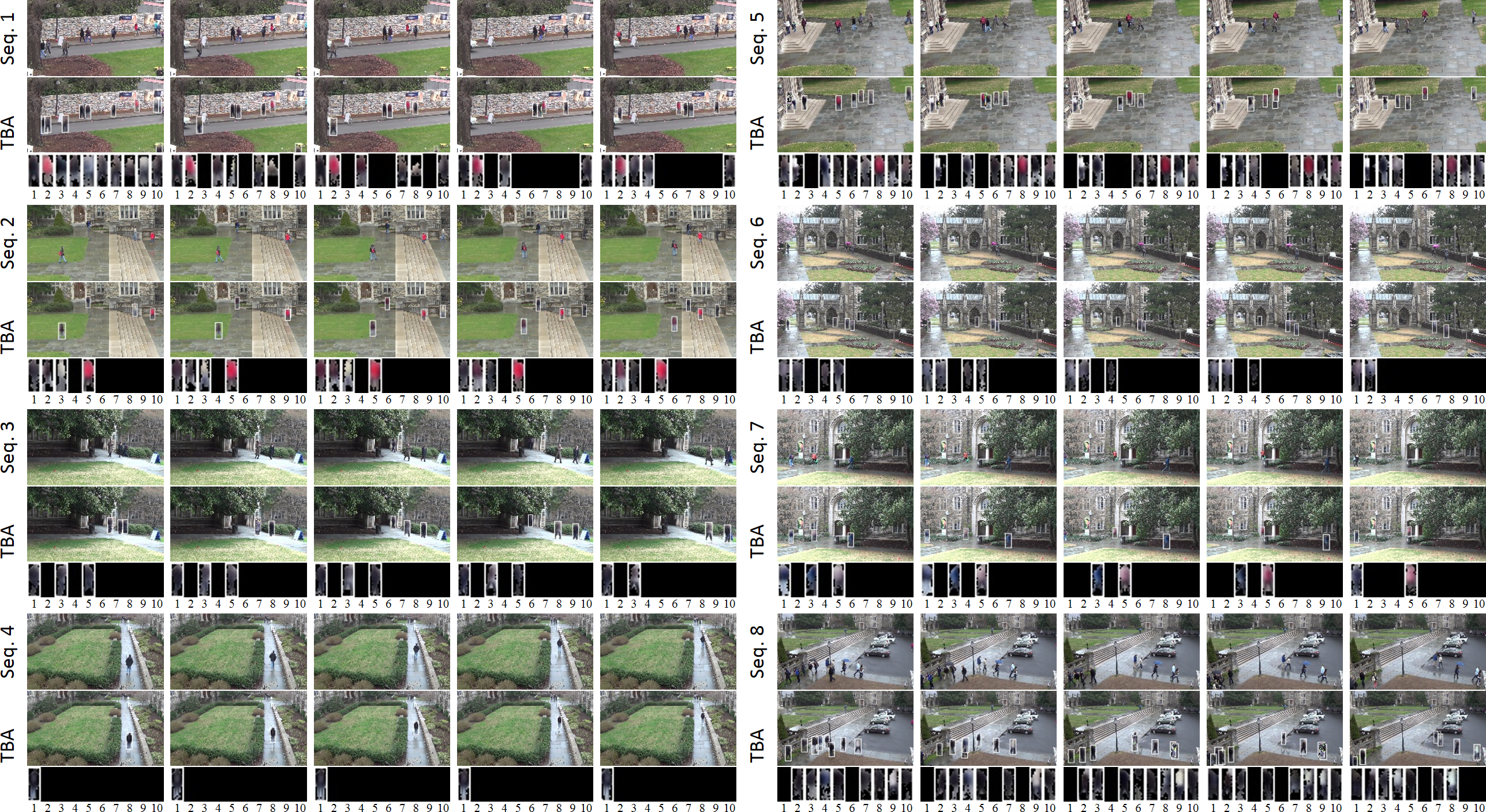}
	\caption{Qualitative results of TBA on DukeMTMC.
		For each sequence, we show the input frames (top), reconstructed frames (middle), and the tracker outputs (bottom). 
		For each frame, tracker outputs from left to right correspond to tracker 1 to $I$ (here $I\!=\!10$), respectively.
		Each tracker output $\mc{Y}_{t,i}$ is visualized as $\lt(y^c_{t,i}~\bm{Y}_{t,i}^s \odot \bm{Y}_{t,i}^a\rt) \in [0,1]^{U\!\times\!V\!\times\!D}$.}
	\label{fig-quali-ped}
	\vspace{9pt}
\end{figure*}

Fig.~\ref{fig-quali-ped} shows sampled qualitative results.
TBA performs well under various situations:
(i)~frequent object appearing/disappearing;
(ii)~highly-varying object numbers, e.g., a single person (Seq. 4) or ten persons (Frame 1 in Seq. 1);
(iii)~frequent object occlusions, e.g., when people walk towards each other (Seq. 1);
(iv)~perspective scale changes, e.g., when people walk close to the camera (Seq. 3);
(v)~frequent shape/appearance changes;
(vi)~similar shapes/appearances for different objects (Seq. 6).

Quantitative performances are presented in Table~\ref{tab-quanti-ped}.
We can see that TBA gains an IDF1 of 82.4\%, a MOTA of 79.6\%, and a MOTP of 80.4\% which is the highest, being very competitive to the state-of-the-art methods in performance.
However, unlike these methods, our model is the first one free of any training labels or extracted features.

\subsection{Visualizing the RAT}

\begin{figure}[H]
	\centering
	\abovecaptionskip=5pt
	\belowcaptionskip=-15pt
	\includegraphics[width=0.47\textwidth]{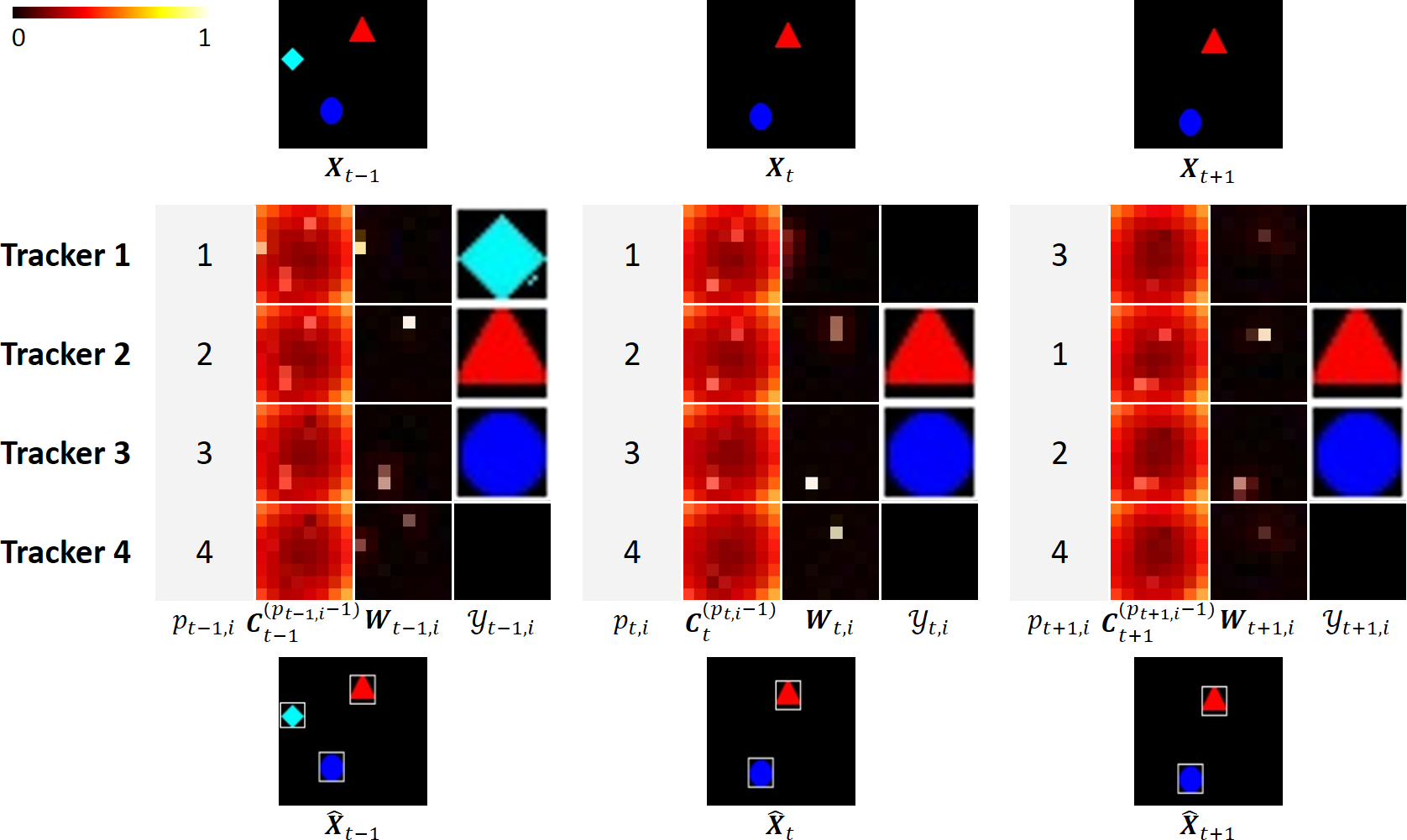}
	\caption{Visualization of the RAT on Sprites-MOT.
		Both the memory $\bm{C}_t$ and the attention weight $\bm{W}_{t,i}$ are visualized as $M\!\times\!N$ ($8\!\times\!8$) matrices, where for $\bm{C}_t$ the matrix denotes its channel mean $\frac{1}{S}\sum_{s=1}^{S}\bm{C}_{t,1:M,1:N,s}$ normalized in $[0, 1]$.}
	\label{fig-vis}
\end{figure}

To get more insights into how the model works, we visualize the process of RAT on Sprites-MOT (see Fig.~\ref{fig-vis}).
At time $t$, Tracker $i$ is updated in the $p_{t,i}$-th iteration, using its attention weight $\bm{W}_{t,i}$ to read from and write to the memory $\bm{C}_t^{(p_{t,i} - 1)}$, obtaining $\bm{C}_t^{(p_{t,i})}$.
We can see that the memory content (bright region) related to the associated object is attentively \emph{erased} (becomes dark) by the write operation, thereby preventing the next tracker from reading it again.
Note that at time $(t\!+\!1)$, Tracker 1 is reprioritized with a priority $p_{t+1,1}\!=\!3$ and thus is updated at the 3-rd iteration, and
the memory value has not been modified in the 3-rd iteration by Tracker~1 at which the iteration is terminated~(since $y^c_{t,1}\!<\!0.5$ and $y^c_{t+1,1}\!<\!0.5$).

\section{Related Work}

\paragraph{Unsupervised Learning for Visual Data Understanding}
There are many works focusing on extracting interpretable representations from visual data using unsupervised learning:
some attempt to find low-level disentangled factors (\cite{kulkarni2015deep,chen2016infogan,rolfe2016discrete} for images and  \cite{ondruvska2016deep,karl2016deep,greff2017neural,denton2017unsupervised,fraccaro2017disentangled} for videos),
some aim to extract mid-level semantics (\cite{le2011learning,moreno2016overcoming,huang2015efficient} for images and \cite{jojic2001learning,winn2005generative,wulff2014modeling,hsieh2018learning} for videos),
while the remaining seek to discover high-level semantics (\cite{eslami2016attend,yan2016perspective,rezende2016unsupervised,stewart2017label,wu2017neural,eslami2018neural} for images and \cite{watters2017visual,wu2017learning} for videos).
However, none of these works deal with MOT tasks.
To the best of our knowledge, the proposed method first achieves unsupervised end-to-end learning of MOT.

\paragraph{Data Association for online MOT}
In MOT tasks, data association can be either offline \cite{zhang2008global,niebles2010efficient,kuo2010multi,berclaz2011multiple,pirsiavash2011globally,butt2013multi,milan2014continuous} or online \cite{turner2014complete,bae2014robust,wu2007detection}, deterministic \cite{perera2006multi,huang2008robust,xing2009multi} or probabilistic \cite{schulz2003people,blackman2004multiple,khan2005mcmc,vo2006gaussian}, greedy \cite{breitenstein2009robust,breitenstein2011online,shu2012part} or global \cite{reilly2010detection,kim2012online,qin2012improving}.
Since the proposed RAT deals with online MOT and uses soft attention to greedily associate data based on tracker confidence ranking, it belongs to the probabilistic and greedy online methods.
However, unlike these traditional methods, RAT is learnable, i.e., the tracker array can learn to generate matching features, evolve tracker states, and modify input features.
Moreover, as RAT is not based on TBD and is end-to-end, the feature extractor can also learn to provide discriminative features to ease data association.

\section{Conclusion}

We introduced the TBA framework which achieves unsupervised end-to-end learning of MOT tasks.
We also introduced the RAT to improve the robustness of data association.
We validated our model on different tasks, showing its potential for real applications such as video surveillance.
Our future work is to extend the model to handle videos with dynamic backgrounds.
We hope our method could pave the way towards more general unsupervised MOT.

\begin{table*}[th]
	\small
	\centering
	\tabcolsep=5pt
	\abovecaptionskip=0pt
	\caption{Tracking performances of different methods on DukeMTMC.}
	\begin{threeparttable}
		\begin{tabular}{ccccccccccccc}
			\toprule
			\tb{Method} &\tb{IDF1\ua}&\tb{IDP\ua}&\tb{IDR\ua}&\tb{MOTA\ua}&\tb{MOTP\ua}&\tb{FAF\da}&\tb{MT\ua}&\tb{ML\da}&\tb{FP\da}&\tb{FN\da}&\tb{IDS\da}&\tb{Frag\da}\\
			\midrule
			DeepCC \cite{ristani2018features} &  89.2  &  91.7  &  86.7  &  87.5  &  77.1  &  0.05  & 1,103 &   29 & 37,280  &  94,399  & 202  &  753 \\
			TAREIDMTMC \cite{jiang2018online} &  83.8  &  87.6  &  80.4  &  83.3  &  75.5  &  0.06  & 1,051 &   17 & 44,691  & 131,220  & 383  & 2,428 \\
			\tb{TBA (ours)}\tnote{*}          &  82.4  &  86.1	&  79.0	 &  79.6  &  80.4  &  0.09  & 1,026 &   46 & 64,002  & 151,483  & 875  & 1,481  \\
			MYTRACKER \cite{yoon2018multiple} &  80.3  &  87.3  &  74.4  &  78.3  &  78.4  &  0.05  &  914  &   72 & 35,580  & 193,253  & 406  & 1,116  \\
			MTMC\_CDSC \cite{tesfaye2017multi}&  77.0  &  87.6  &  68.6  &  70.9  &  75.8  &  0.05  &  740  &  110 & 38,655  & 268,398  & 693  & 4,717  \\
			PT\_BIPCC \cite{maksai2017non}    &  71.2  &  84.8  &  61.4  &  59.3  &  78.7  &  0.09  &  666  &  234 & 68,634  & 361,589  & 290  &   783  \\
			BIPCC \cite{ristani2016performance}&  70.1 &  83.6  &  60.4  &  59.4  &  78.7  &  0.09  &  665  &  234 & 68,147  & 361,672  & 300  &   801  \\
			\bottomrule
		\end{tabular}
		\begin{tablenotes}
			\item[*] The results are hosted at \url{https://motchallenge.net/results/DukeMTMCT}, where our TBA tracker is named as `MOT\_TBA'.
		\end{tablenotes}
		\vspace{-5pt}
	\end{threeparttable}\label{tab-quanti-ped}
\end{table*}

{\small
\bibliographystyle{ieee_fullname}
\bibliography{tba}
}

\clearpage
\appendix

\section{Supplementary Materials for Experiments}

\subsection{Implementation Details}\label{app_implement}

\paragraph{Model Configuration}
There are some common model configurations for all tasks.
For the $\op{NN}^{feat}$ defined in (\ref{eq_feat}), we set it as a FCN, where each convolution layer is composed via convolution, adaptive max-pooling, and ReLU and the convolution stride is set to 1 for all layers.
For the $\op{RNN}$ defined in (\ref{eq_read3}), we set it as a Gated Recurrent Unit (GRU) \cite{cho2014learning} to capture long-range temporal dependencies.
For the $\op{NN}^{out}$ defined in (\ref{eq_out}), we set it as a Fully-Connected network (FC), where the ReLU is chosen as the activation function for each hidden layer.
For the loss defined in (\ref{eq_loss}), we set $\lambda\!=\!1$.
For the model configurations specified to each task, please see in Table~\ref{tab-config}.
Note that to use attention, the receptive field of $\bm{c}_{t,m,n}$ is crafted as a local region on $\bm{X}_t$, i.e., 40$\times$40 for MNIST-MOT and Sprites-MOT, and 44$\times$24 for DukeMTMC (this can be calculated using the FCN hyper-parameters in Table~\ref{tab-config}).

\paragraph{Training Configuration}
For MNIST-MOT and Sprites-MOT, we split the data into a proportion of 90/5/5 for training/validation/test;
for DukeMTMC, we split the provided training data into a proportion of 95/5 for training/validation.
For all tasks, in each iteration we feed the model with a mini-batch of
64 subsequences of length 20. 
During the forward pass, the tracker states and confidences at the last time step are preserved to initialize the next iteration.
To train the model, we minimize the averaged loss on the training set w.r.t. all model parameters $\bm{\Theta}\!=\!\{\bm{\theta}^{feat}, \bm{\theta}^{upd}, \bm{\theta}^{out}\}$ using Adam \cite{kingma2014adam} with a learning rate of $5\!\times\!10^{-4}$.
Early stopping is used to terminate training.

\begin{table*}[t]
	\small
	\centering
	\tabcolsep=8pt
	\abovecaptionskip=0pt
	\caption{Model configurations specified to each task, where `conv h$\times$w' denotes a convolution layer with kernel size h$\times$w, `fc' denotes a fully-connected layer, and `out' denotes an output layer. Note that for $\op{NN}^{feat}$, the first layer has two additional channels than $\bm{X}_t$, which are the 2D image coordinates (as mentioned in Sec.~\ref{subsec_att}).}
	\begin{tabular}{crlrlrl}
		\toprule
		\tb{Hyper-parameter} & \multicolumn{2}{c}{\tb{MNIST-MOT}}   & \multicolumn{2}{c}{\tb{Sprites-MOT}}    & \multicolumn{2}{c}{\tb{DukeMTMC}}       \\
		\toprule
		Size of $\bm{X}_t$: $[H, W, D]$       & \multicolumn{2}{c}{[128, 128, 1]} & \multicolumn{2}{c}{[128, 128, 3]}  & \multicolumn{2}{c}{[108, 192, 3]}  \\
		\midrule
		Size of $\bm{C}_t$: $[M, N, S]$       & \multicolumn{2}{c}{[8, 8, 50]}    & \multicolumn{2}{c}{[8, 8, 20]}     & \multicolumn{2}{c}{[9, 16, 200]}  \\
		\midrule
		Size of $\bm{Y}^a_{t,i}$: $[U, V, D]$ & \multicolumn{2}{c}{[28, 28, 1]}   & \multicolumn{2}{c}{[21, 21, 3]}    & \multicolumn{2}{c}{[9, 23, 3]}  \\
		\midrule
		Size of $\bm{h}_{t,i}$: $R$           & \multicolumn{2}{c}{200}           & \multicolumn{2}{c}{80}             & \multicolumn{2}{c}{800}  \\
		\midrule
		Tracker number: $I$                   & \multicolumn{2}{c}{4}             & \multicolumn{2}{c}{4}              & \multicolumn{2}{c}{10}  \\
		\midrule
		Layer number: $K$                     & \multicolumn{2}{c}{1}             & \multicolumn{2}{c}{3}              & \multicolumn{2}{c}{3}  \\
		\midrule
		Coef. of $[\widehat{s}^x_{t,i}, \widehat{s}^y_{t,i}]$:     $[\eta^{x}, \eta^{y}]$      & \multicolumn{2}{c}{[0, 0]}     & \multicolumn{2}{c}{[0.2, 0.2]} & \multicolumn{2}{c}{[0.4, 0.4]}\\
		\midrule
		\multirow{5}[3]{*}{Layer sizes of $\op{NN}^{feat}$ (FCN)} 
		& [128, 128, 3] & (conv 5$\times$5)     & [128, 128, 5] & (conv 5$\times$5)      & [108, 192, 5]  & (conv 5$\times$5)   \\
		& [64, 64, 32]  & (conv 3$\times$3)     & [64, 64, 32]  & (conv 3$\times$3)      & [108, 192, 32] & (conv 5$\times$3)   \\
		& [32, 32, 64]  & (conv 1$\times$1)     & [32, 32, 64]  & (conv 1$\times$1)      & [36, 64, 128]  & (conv 5$\times$3)   \\
		& [16, 16, 128] & (conv 3$\times$3)     & [16, 16, 128] & (conv 3$\times$3)      & [18, 32, 256]  & (conv 3$\times$1)   \\
		& [8, 8, 256]   & (conv 1$\times$1)     & [8, 8, 256]   & (conv 1$\times$1)      & [9, 16, 512]   & (conv 1$\times$1)   \\
		& [8, 8, 50]    & (out)                 & [8, 8, 20]    & (out)                  & [9, 16, 200]   & (out)            \\
		\midrule
		\multirow{3}[0]{*}{Layer sizes of $\op{NN}^{out}$ (FC)} 
		& 200            & (fc)                  & 80            & (fc)                  & 800            & (fc)          \\
		& 397            & (fc)                  & 377           & (fc)                  & 818            & (fc)         \\
		& 787            & (out)                 & 1772          & (out)                 & 836            & (out)         \\
		\toprule
		Number of parameters                   & \multicolumn{2}{c}{1.21 M}        & \multicolumn{2}{c}{1.02 M}         & \multicolumn{2}{c}{5.65 M}  \\
		\bottomrule
	\end{tabular}\label{tab-config}
\end{table*}

\subsection{MNIST-MOT}\label{app_mnist}

As a pilot experiment, we focus on testing whether our model can robustly track the position and appearance of each object that can appear/disappear from the scene.
Thus, we create a new MNIST-MOT dataset
containing 2M frames,
where each frame is of size 128$\times$128$\times$1,
consisting of a black background and at most three moving digits. 
Each digit is a 28$\times$28$\times$1 image patch
randomly drawn from the MNIST dataset \cite{lecun1998gradient}, 
moves towards a random direction, and appears/disappears only once. 
When digits overlap, pixel values are added and clamped in $[0,1]$.
To solve this task, for TBA configurations 
we set the tracker number $I\!=\!4$ and layer number $K\!=\!1$, and 
fix the scale $s_{t,i}^{x}\!=s_{t,i}^{y}\!=1$ and shape $\bm{Y}_{t,i}^{s}\!=\!\bm{1}$, thereby 
only compositing a single layer by adding up all transformed appearances. 
We also clamp the pixel values of the reconstructed frames in $[0,1]$ for all configurations.

\begin{figure}[t]
	\centering
	\includegraphics[width=0.4\textwidth]{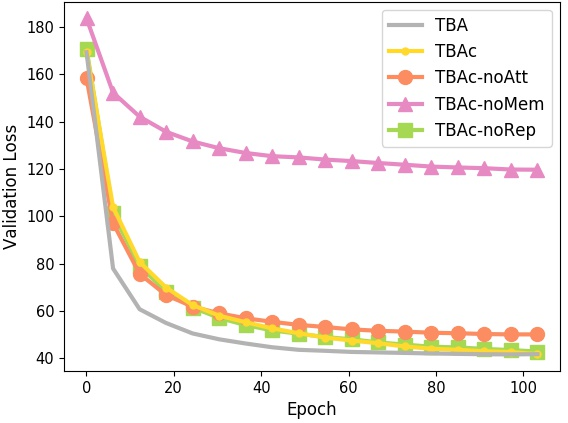}
	\caption{Training curves of different configurations on MNIST-MOT.}
	\label{fig-curve-mnist}
\end{figure}

Training curves are shown in Fig.~\ref{fig-curve-mnist}.
The TBA, TBAc, and TBAc-noRep have similar validation losses which are slightly better than that of TBAc-noAtt.
Similar to the results on Sprites-MOT, TBA converges the fastest, and TBAc-noMem has a significantly higher validation loss as all trackers are likely to focus on a same object, which affects the reconstruction.

\begin{figure*}[t]
	\centering
	\includegraphics[width=0.994\textwidth]{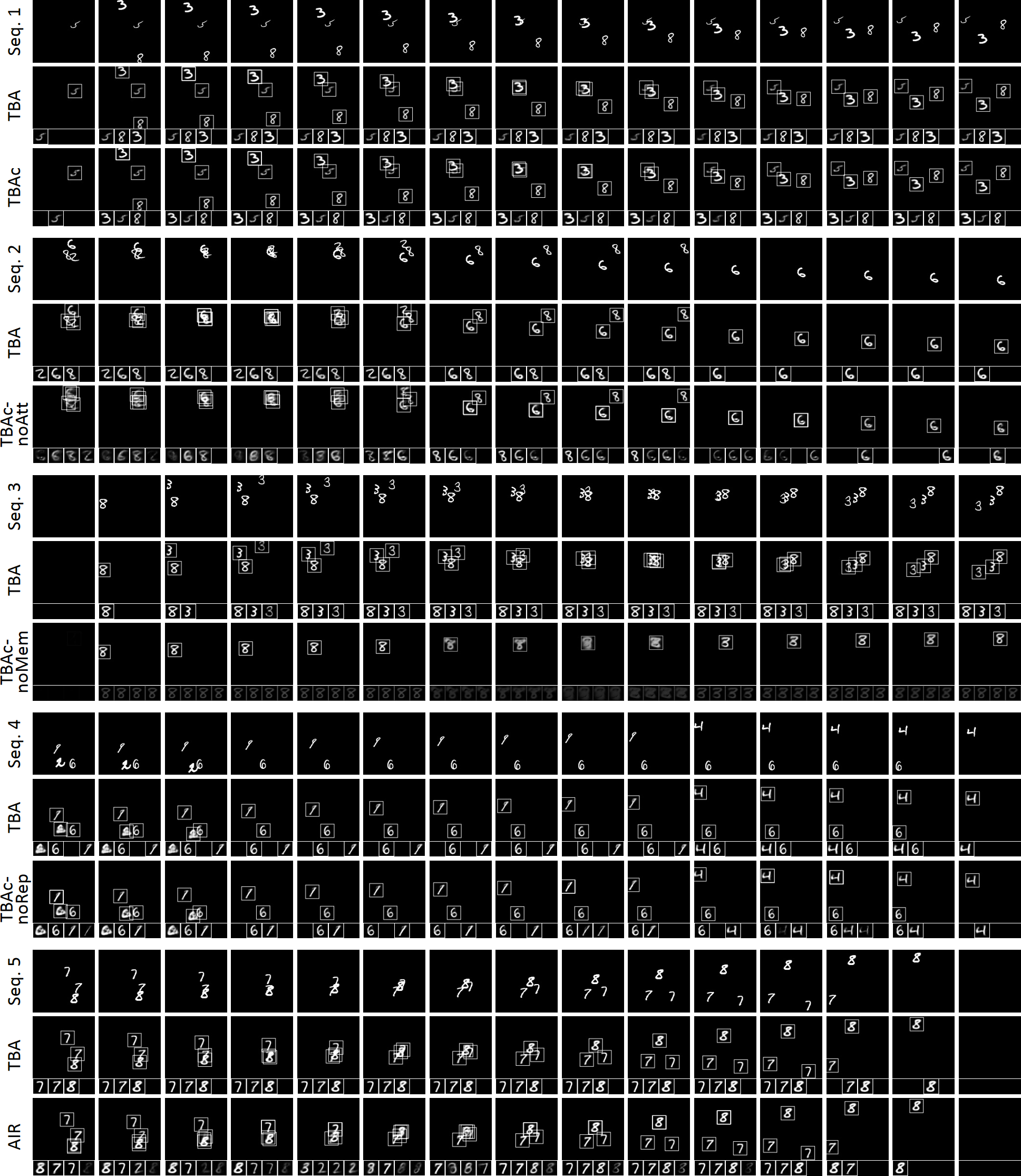}
	\caption{Qualitative results of different configurations on MNIST-MOT.
		For each configuration, we show the reconstructed frames (top) and the tracker outputs (bottom). 
        For each frame, tracker outputs from left to right correspond to tracker 1 to $I$ (here $I\!=\!4$), respectively.
        Each tracker output $\mc{Y}_{t,i}$ is visualized as $\lt(y^c_{t,i}~\bm{Y}_{t,i}^s \odot \bm{Y}_{t,i}^a\rt) \in [0,1]^{U\!\times\!V\!\times\!D}$.}
	\label{fig-quali-mnist}
\end{figure*}

Qualitative results are shown in Fig.~\ref{fig-quali-mnist}.
Similar phenomena are observed as in Sprites-MOT, revealing the importance of the disabled mechanisms.
Specifically, as temporal dependency is not considered in AIR, overlapped objects are failed to be disambiguated (Seq. 5).

We further quantitatively evaluate different configurations.
Results are reported in Table \ref{tab-quanti-mnist}, which are similar to those of the Sprites-MOT.

\begin{table*}[t]
	\small
	\centering
	\tabcolsep=5pt
	\abovecaptionskip=0pt
	\caption{Tracking performances of different configurations on MNIST-MOT.}
	\begin{tabular}{ccccccccccccc}
		\toprule
		\tb{Configuration} &\tb{IDF1\ua}&\tb{IDP\ua}&\tb{IDR\ua}&\tb{MOTA\ua}&\tb{MOTP\ua}&\tb{FAF\da}&\tb{MT\ua}&\tb{ML\da}&\tb{FP\da}&\tb{FN\da}&\tb{IDS\da}&\tb{Frag\da}\\
		\midrule
		TBA         &  99.6  &  99.6  &  99.6  &  99.5  &  78.4  &     0  &  978  &    0  &    49  &     49  &     22  &     7  \\
		TBAc        &  99.2  &  99.3  &  99.2  &  99.4  &  78.1  &  0.01  &  977  &    0  &    54  &     52  &     26  &    11  \\
		TBAc-noAtt  &  45.2  &  43.9  &  46.6  &  59.8  &  81.8  &  0.20  &  976  &    0  & 1,951  &    219  &  6,762  &    86  \\
		TBAc-noMem  &     0  &    --  &     0  &     0  &    --  &     0  &    0  &  983  &     0  & 22,219  &      0  &     0  \\
		TBAc-noRep  &  94.3  &  92.9  &  95.7  &  98.7  &  77.8  &  0.01  &  980  &    0  &   126  &     55  &    103  &    10  \\
		\bottomrule
	\end{tabular}\label{tab-quanti-mnist}
\end{table*}

\end{document}